\documentclass{article}

\usepackage{cite}
\usepackage{arxiv}

\usepackage[utf8]{inputenc} 
\usepackage[T1]{fontenc}    
\usepackage{hyperref}       
\usepackage{url}            
\usepackage{booktabs}       
\usepackage{amsfonts}       
\usepackage{nicefrac}       
\usepackage{microtype}      
\usepackage{lipsum}
\usepackage{graphicx}

\title{A Deep Image Compression Framework for Face Recognition}

\author{
  Nai Bian \\
  School of Microelectronics\\
  Xi'an Jiaotong University\\
  Xi'an, China \\
  \texttt{biannai120@stu.xjtu.edu.cn} \\
   \And
  Feng Liang \\
  School of Microelectronics\\
  Xi'an Jiaotong University\\
  Xi'an, China \\
  \texttt{fengliang@xjtu.edu.cn} \\
  \AND
  Haisheng Fu \\
  School of Microelectronics \\
  Xi'an Jiaotong University \\
  Xi'an, China \\
  \texttt{fhs4118005070@stu.xjtu.edu.cn} \\
  \And
  Bo Lei \\
  School of Microelectronics \\
  Xi'an Jiaotong University \\
  Xi'an, China \\
  \texttt{lei\_bo@stu.xjtu.edu.cn} \\
}

\begin{document}
\maketitle

\begin{abstract}
Face recognition technology has advanced rapidly and has been widely used in various applications. Due to the extremely huge amount of data of face images and the large computing resources required correspondingly in large-scale face recognition tasks, there is a requirement for a face image compression approach that is highly suitable for face recognition tasks. In this paper, we propose a deep convolutional autoencoder compression network for face recognition tasks. In the compression process, deep features are extracted from the original image by the convolutional neural networks to produce a compact representation of the original image, which is then encoded and saved by existing codec such as PNG. This compact representation is utilized by the reconstruction network to generate a reconstructed image of the original one. In order to improve the face recognition accuracy when the compression framework is used in a face recognition system, we combine this compression framework with a existing face recognition network for joint optimization. We test the proposed scheme and find that after joint training, the Labeled Faces in the Wild (LFW) dataset compressed by our compression framework has higher face verification accuracy than that compressed by JPEG2000, and is much higher than that compressed by JPEG.
\end{abstract}


\section{Introduction}
As a prominent representative of the well developed deep learning technology, face recognition has been successfully used in tasks of different scales. Face identification on mobile phones is a common example of small-scale face recognition tasks while urban intelligent monitoring system is an example of large-scale ones. Large-scale face recognition tasks like those systems analysis the images captured by cameras in various places to identify faces. In this process, the system needs to compare the obtained image with hundreds of millions of face images in its database. Such a large number of face images not only occupy a large amount of storage resources, but also expend a large amount of computing resources. So it is of great significance to propose an efficient image compression approach that can be adapted to tasks like face recognition and image classification.

\begin{figure}[htbp]
\centering
\includegraphics[width=6cm]{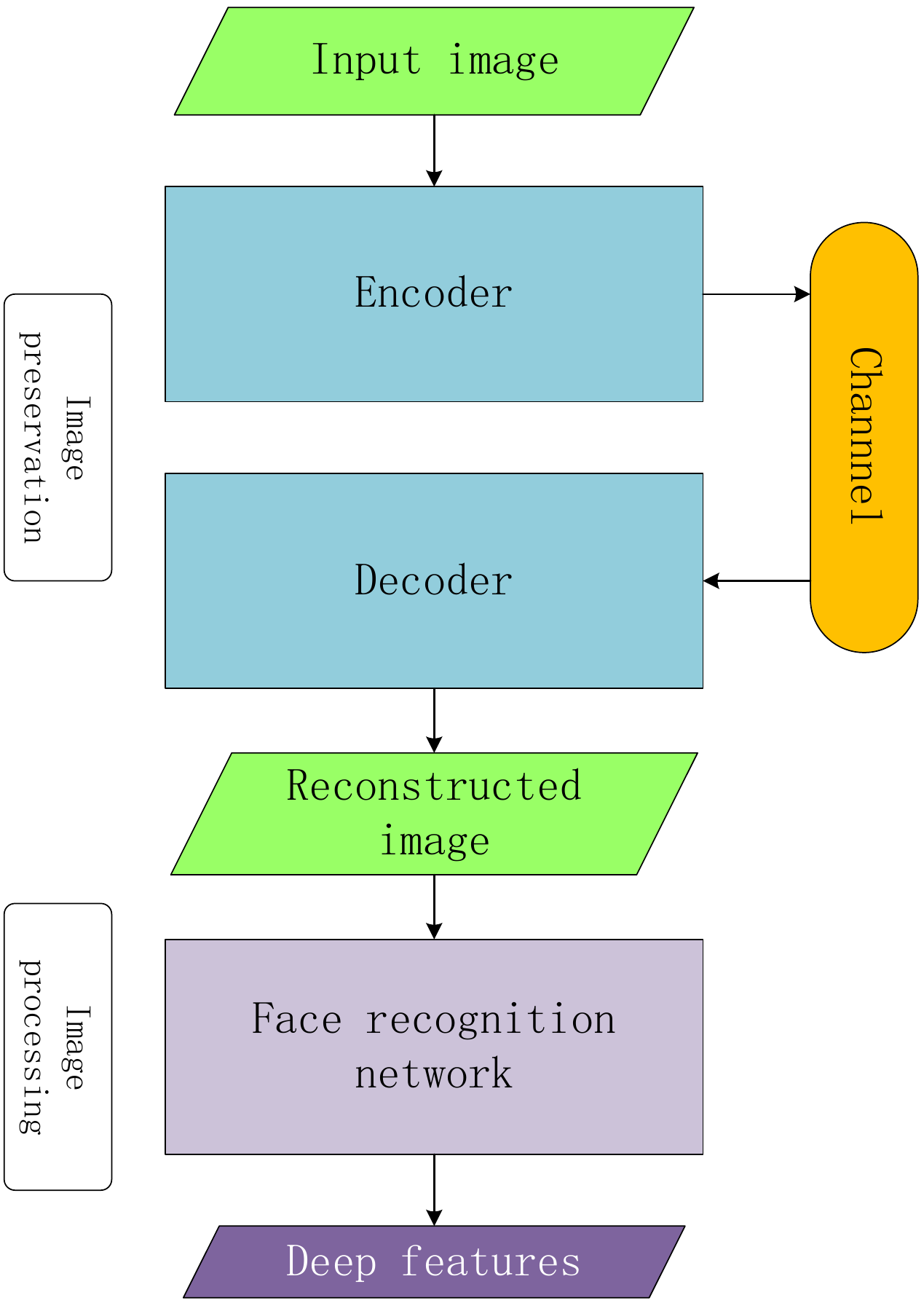}
\caption{The components of face recognition system.}
\label{fig1}
\end{figure}

\begin{figure}[htbp]
\centering
\includegraphics[width=12cm]{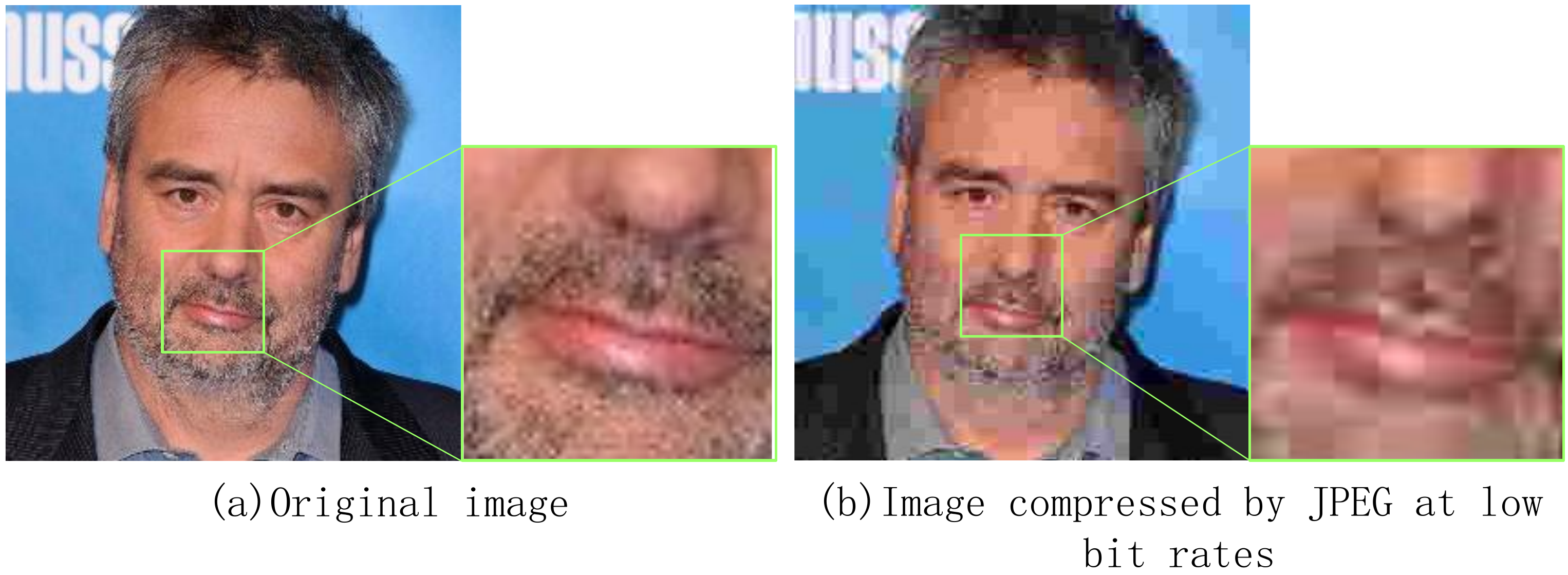}
\caption{Blocking artifacts of images compressed by JPEG at low bit rates.}
\label{fig2}
\end{figure}

\begin{figure}[htbp]
\centering
\includegraphics[width=8.4cm]{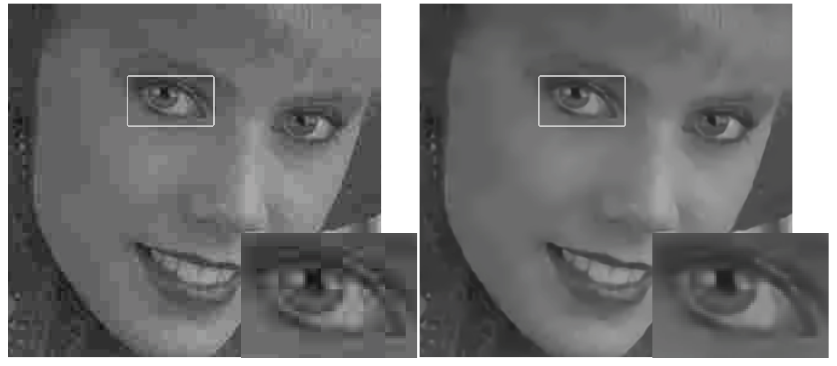}
\caption{Example JPEG compressed image with blocking artifacts and the restored image by AR-CNN \cite{dong2015compression}.}
\label{fig10}
\end{figure}

As shown in Figure~\ref{fig1}, the process of a face recognition system includes two parts: preservation and processing of face images. The original image data are compressed, encoded, decoded and restored by the image codec. Then the reconstructed image is processed by the face recognition network to obtain the deep features. The deep features here are often in the form of first order tensors. The basic work of face recognition is to compare the difference between the deep features of two face images.
Face images in many face recognition applications are compressed and reconstructed by conventional schemes such as JPEG. But there would be some annoying blocking artifacts in a picture when compressed at a low bit rates by JPEG, which is presented in Figure~\ref{fig2}. Such artifacts not only affect image quality, but also affect the performance of feature extraction in subsequent image processing tasks such as face recognition and image classification.
Some post-processing method such as AR-CNN \cite{dong2015compression} (Artifacts Reduction Convolutional Neural Network) is indeed useful for deblocking and arifacts reduction, which is shown in Figure~\ref{fig10}. However, these deblocking networks uaually involve an iterative process that consumes high computational resources. Moreover, the optimization goal of the deblocking network is not entirely consistent with that of the face recognition network. Training a network that enhances image quality alone does not necessarily mean achieving the optimal result of a face recognition task.

Recent years, there has been many learning-based image compression approaches showing excellent performance and some methods \cite{agustsson2017soft,agustsson2018generative,theis2017lossy,toderici2017full} even perform better than JPEG, JPEG2000 and BPG (a state of the art lossy coding method). Besides, due to its neural network structure and characteristic of backpropagation, a learing-based approach is possible to be combined with another face recognition network and trained jointly.
Inspired by \cite{jiang2017end}, in this paper, we use a compression-reconstruction network in the form of autoencoder combined with a standard image lossless codec PNG to replace the conventional codec in Figure~\ref{fig1}. Aiming at improving the accuracy of face recognition, the compression-reconstruction network and face recognition network are trained as a whole. Finally, a higher verification accuracy was obtained on the LFW face dataset compressed by our jointly optimized compression network than on the LFW face dataset compressed by JPEG and JPEG2000.

The sections in this paper are organized as follows. Section 2 introduces the  compression-reconstruction-recognition joint framework proposed in this paper, and describes their network structures and training strategies. Section 3 evaluates the effect of our proposed compression scheme and compares and analyzes the face verification accuracy on LFW dataset compressed by our compression network, JPEG and JPEG2000.

\section{CNN based face image compression-reconstruction-recognition joint framework}

\subsection{The overall structure of the joint framework}

\begin{figure}[htbp]
\centerline{\includegraphics[width=10.26cm]{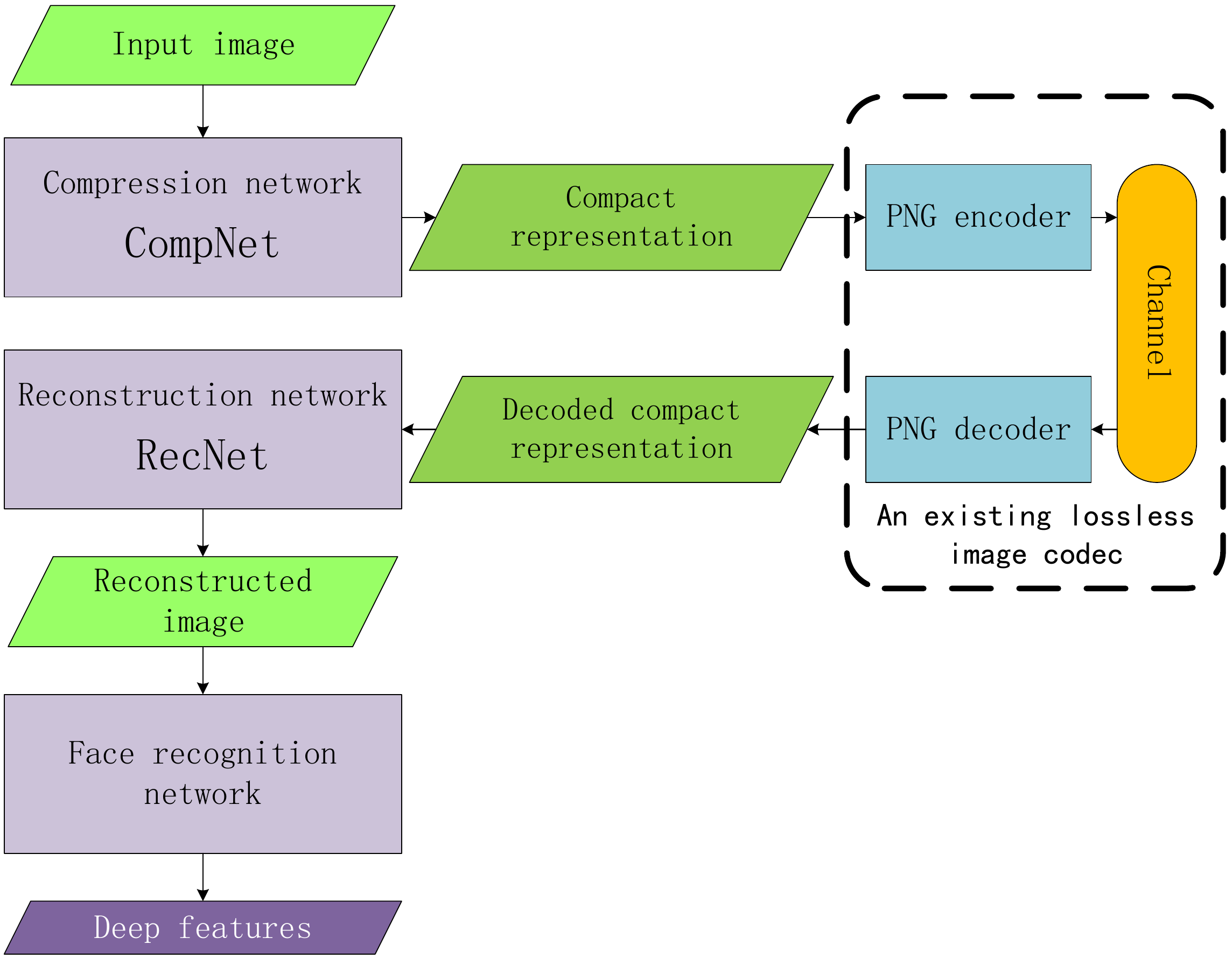}}
\caption{The overall structure of compression-reconstruction-recognition franmework.}
\label{fig3}
\end{figure}

The overall structure of the proposed compression-reconstruction-recognition framework is shown in Figure~\ref{fig3}. We replace the conventional image compression codec in the face recognition system with a compression network composed of two convolutional neural networks called CompNet and RecNet and a standard lossless compression codec (such as PNG). They are combined to achieve the purpose of face image compression and recovery. We directly connect this compression-reconstruction framework with a face recognition network.

In this system, a face image is input into the compression network CompNet to produce a compact representation of the original image. The compact representation is 1/8 of the size of the original image and is encoded and decoded with PNG (a standard lossless image codec). This compact representation is then sent to the reconstruction network RecNet to reconstruct the original image. The reconstructed image passes through a face recognition network to generate a series of feature vectors, which is called the deep features of the face. These feature vectors can be used to implement face recognition through a multi-classifier such as LMCL \cite{wang2018cosface} (Large Margin Cosine Loss, the classifier and loss function used in this paper), or directly compare with that of another face image.
We jointly train the CompNet and the RecNet, so that the compact representation produced by CompNet can preserved the useful information of the original image as much as possible. Reconstructed image generated by RecNet can be restored as close as possible to the original image. We then connect the proposed compression-reconstruction framework and another existing face recognition network and jointly train it with the pre-trained model parameters of both networks for initialization. We assume that the parameters of the combined network will converge faster than the fully retrained one considering the idea of transfer learning.

\subsection{Design of CompNet and RecNet}

\subsubsection{Network structure of CompNet and RecNet}
\begin{figure}[htbp]
\centerline{\includegraphics[width=10.8cm]{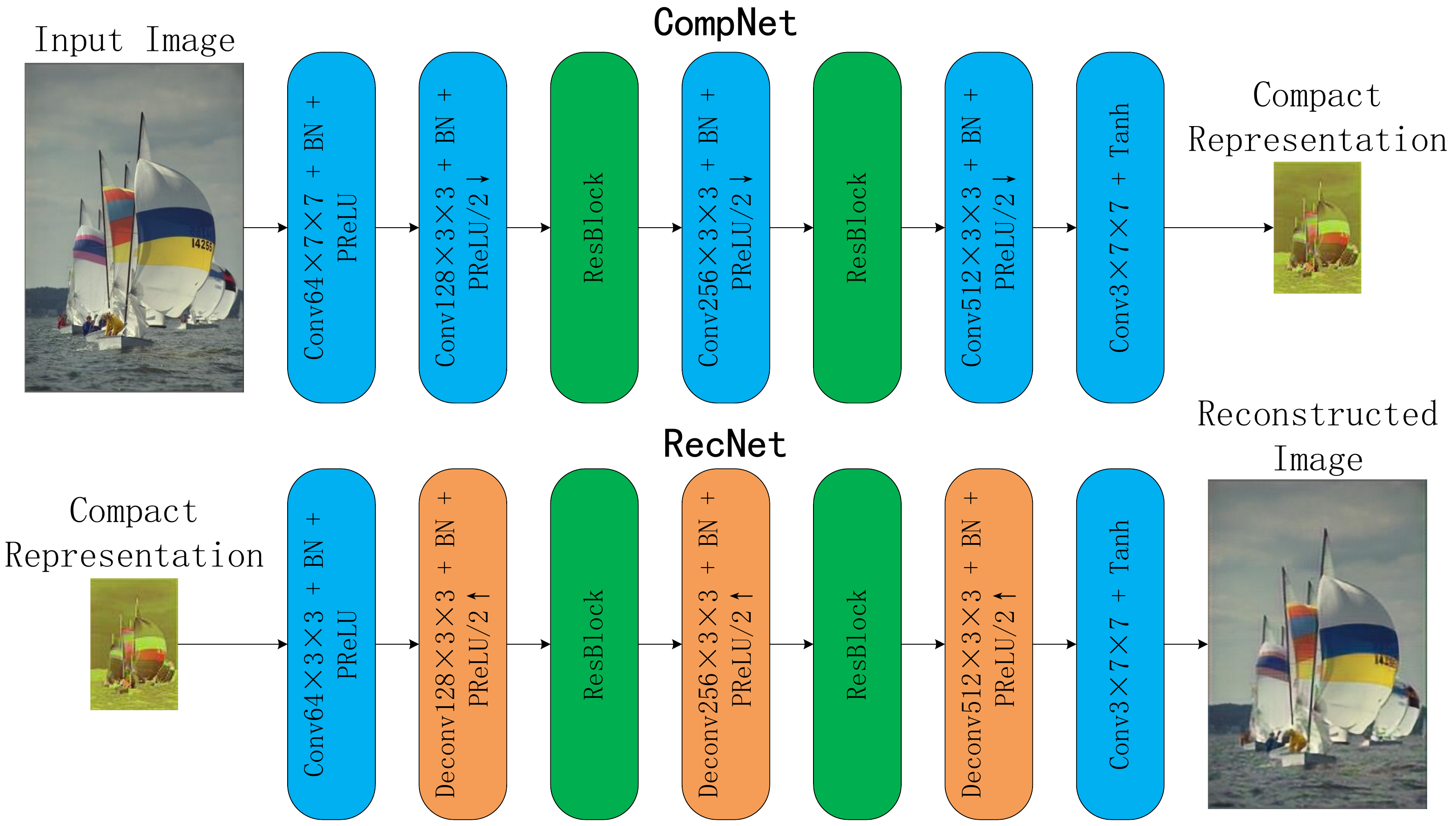}}
\caption{The structure of CompNet and RecNet.}
\label{fig4}
\end{figure}

The architectures of CompNet and RecNet is presented in Figure~\ref{fig4}. It is designed in the form of convolutional autoencoders.

CompNet and RecNet are symmetrical designed. As shown in Figure~\ref{fig4}, CompNet uses a five-layer convolution structure. The second, third, and fourth convolutional layers are called downsampling layers, whose strides are set to 2. Each time a downsampling layer is passed, the height and width of the feature maps become 1/2 of the original one. There are 3 downsampling layers, so the height and width of the final compact map are both 1/8 of the original image.

Correspondingly, there are 3 upsampling layers using deconvolution structure on the RecNet side. Through every upsampling layer, the height and width of feature maps will double, so that a original-sized image can be restored from a compact map of 1/8 of its height and width.

\begin{figure}[htbp]
\centerline{\includegraphics[width=9.6cm]{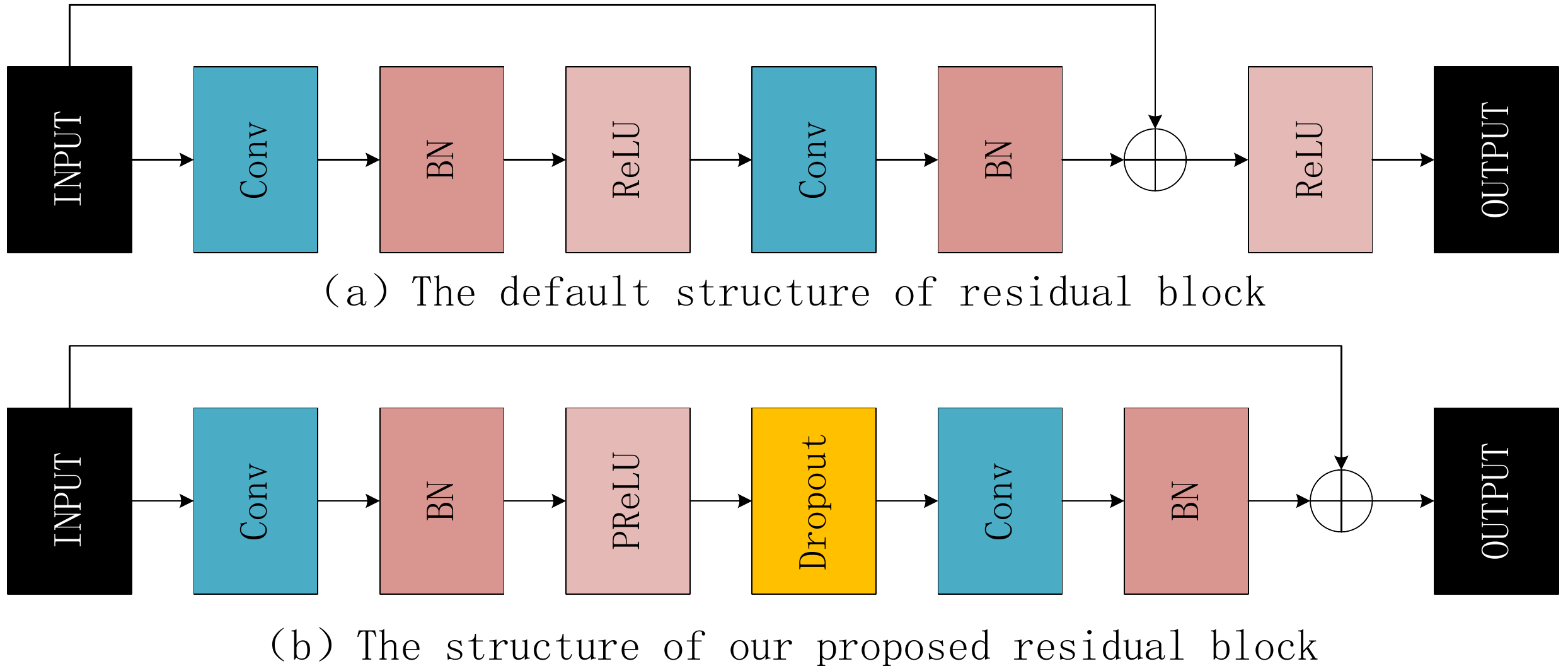}}
\caption{The default structure of residual block \cite{he2016deep} and the structure of the proposed residual block.}
\label{fig5}
\end{figure}

We add a residual block between every two upsampling layers and every two downsampling layers. In addition, inspired by \cite{Liu2018Deep}, we changed the structure of the residual block and added the dropout module to get better results. The structure of the residual block is shown in Figure~\ref{fig5}. In order to achieve faster convergence speed, we replace the default Rectified Linear Unit(ReLU) with a Parametric Rectified Linear Unit(PReLU).

\subsubsection{Quantization and entropy coding}
The values of the original image are integers in [0, 255]. In order to train the network better, all image data will be normalized to floating-point number in [-1, 1] before input into the network. The values of the compact map are still floating-point numbers in [-1, 1]. In order to encode the compact representation whose values range in [-1, 1] with existing codecs, it is necessary to first map those values to [0, 255] and quantify them:
\begin{equation}
    X_{Q}=round(\frac{(X_{E}+1)\times 255}{2})
\end{equation}
where $X_{E}\in\left[1,1\right]$ represents the compact map produced by CompNet, $X_{Q}\in \left [ 0,255 \right ]$ represents the image after quantization.

In practical applications, we use a common standard lossless codec PNG to encode the output of CompNet.
In RecNet, we first need to convert the values of the compact map to floating-point number in [-1, 1] again. That is, $(X_{Q}\times 2/255-1)$ is input into RecNet to obtain a reconstructed image.
Because quantization is a non-differentiable operation, and the gradient cannot be calculated during backpropagation, so the quantization part cannot be directly put into the network during training process. But let's assume that the quantization operation only introduces a negligible error, so we skip the quantization function in training.

\subsubsection{Training of CompNet and RecNet}
We use the open source dataset published by the Computer Vision Lab of ETH Zurich as the training set for our CompNet and RecNet. The reason why we didn't use a face image dataset as training set is that there are few high-definition face datasets, but there is a high requirement on the image definition to train an image compression network. To better train our network, we use several data enhancement methods, such as rotation and scaling. We extract 81,650 image blocks from 1633 high-quality images. These images are losslessly saved as PNG format to avoid compression artifacts. We only use MSE (mean square error) as the loss function of the training network:
\begin{equation}
L=\frac{1}{N}\left | Y_{n}-X_{n}\right
|^{2}
\end{equation}
where $N$ represents batchsize, $X_{n}$ represents the original image, $Y_{n}$ represents the reconstructed image which is the output of RecNet.

CompNet and RecNet are trained together using the SGD (Stochastic Gradient Descent) as the optimizer. We set batchsize to 20 and train the network for 40 epochs with a fixed learning rate of 0.0001.
\subsection{Selection and pre-training of face recognition network}
In recent years, convolutional neural networks based face recognition approaches have well developed, and some \cite{wang2017normface,liu2017sphereface,wang2018cosface,deng2019arcface} have achieved more than 99\% verification accuracy on the LFW dataset utilizing different loss functions. Given the high efficiency of the existing face recognition works, we directly utilize one of these sophisticated solutions and add it to our compression-reconstruction framework. In this work, we use the scheme proposed in Cosface \cite{wang2018cosface}. It uses the sphere20 network structure, and adopts LMCL (Large Margin Cosine Loss) as its loss function. It was pre-trained on CASIA-WebFace dataset, and can achieve an verification accuracy of 99.28\% on the LFW dataset.

\subsubsection{Structure of Cosface Face Recognition Model}
\begin{figure}[htbp]
\centerline{\includegraphics[width=12.9cm]{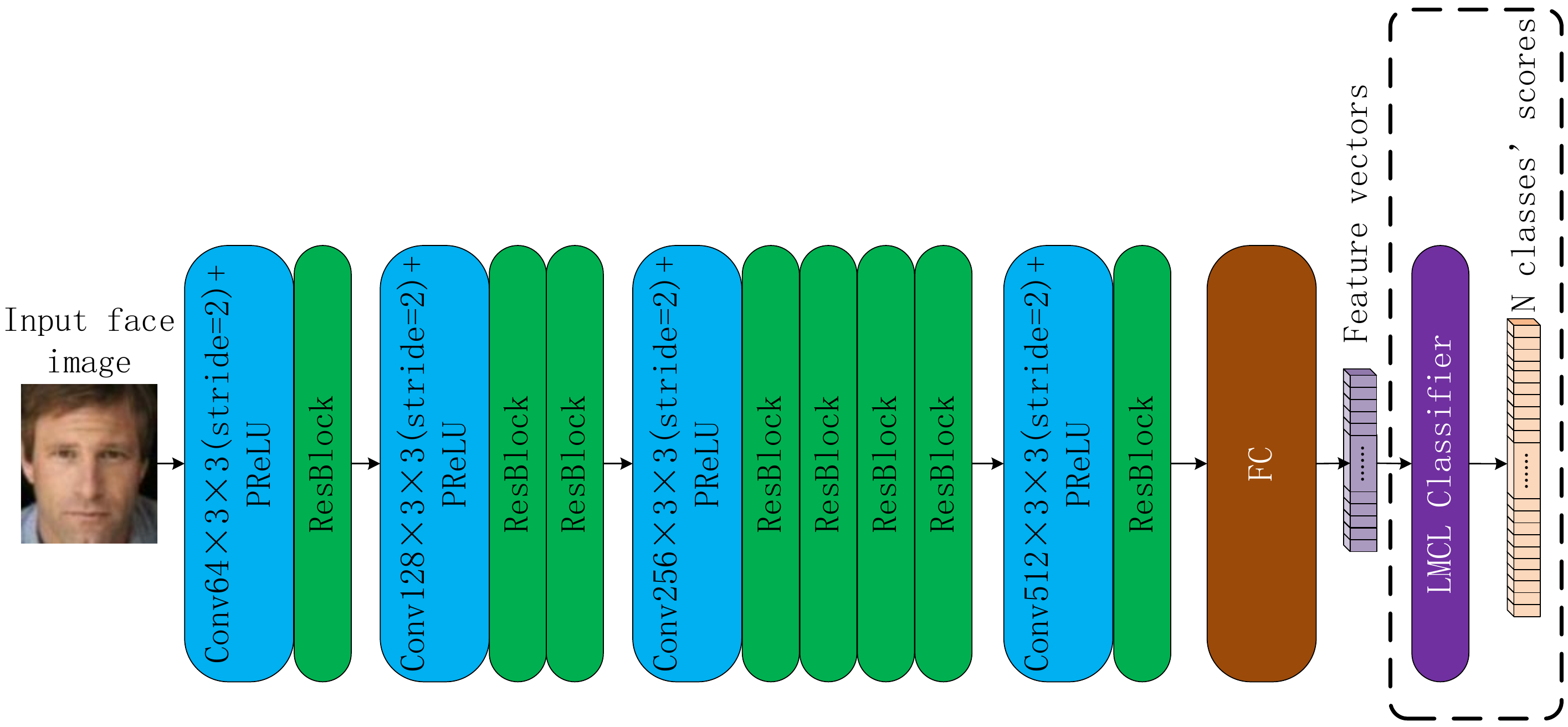}}
\caption{The sphere20 network of Cosface.}
\label{fig6}
\end{figure}

Cosface's Sphere20 networks are shown in Figure~\ref{fig6}. The pre-trained model parameters of the feature extraction part which is outside the dotted box in Figure~\ref{fig6} have been given. The parameters of the classifier part which is inside the dotted box in Figure~\ref{fig6} have to be retrained.

The pre-trained model parameters of the face feature extraction part are imported and frozen during training process. This part does not participate in backpropagation. Only the parameters of the classifier are updated in backpropagation. 452,960 face images from 10,575 individuals in CASIA-WebFace dataset are cropped by MTCNN \cite{zhang2016joint} (Multi-task Cascaded Convolutional Networks, a face detection and calibration framework) to the size of 112$\times$96 and then are used as the training set. Batchsize is set to 100. The gradient descent algorithm is SGD. Learning rate is initially set to 0.1 and gradually reduced to 0.0000001. After 6 epochs of training, we obtained a pre-trained model of the classifier of the Cosface network.

\subsection{Joint training of the combined network}
\begin{figure}[htbp]
\centerline{\includegraphics[width=10.5cm]{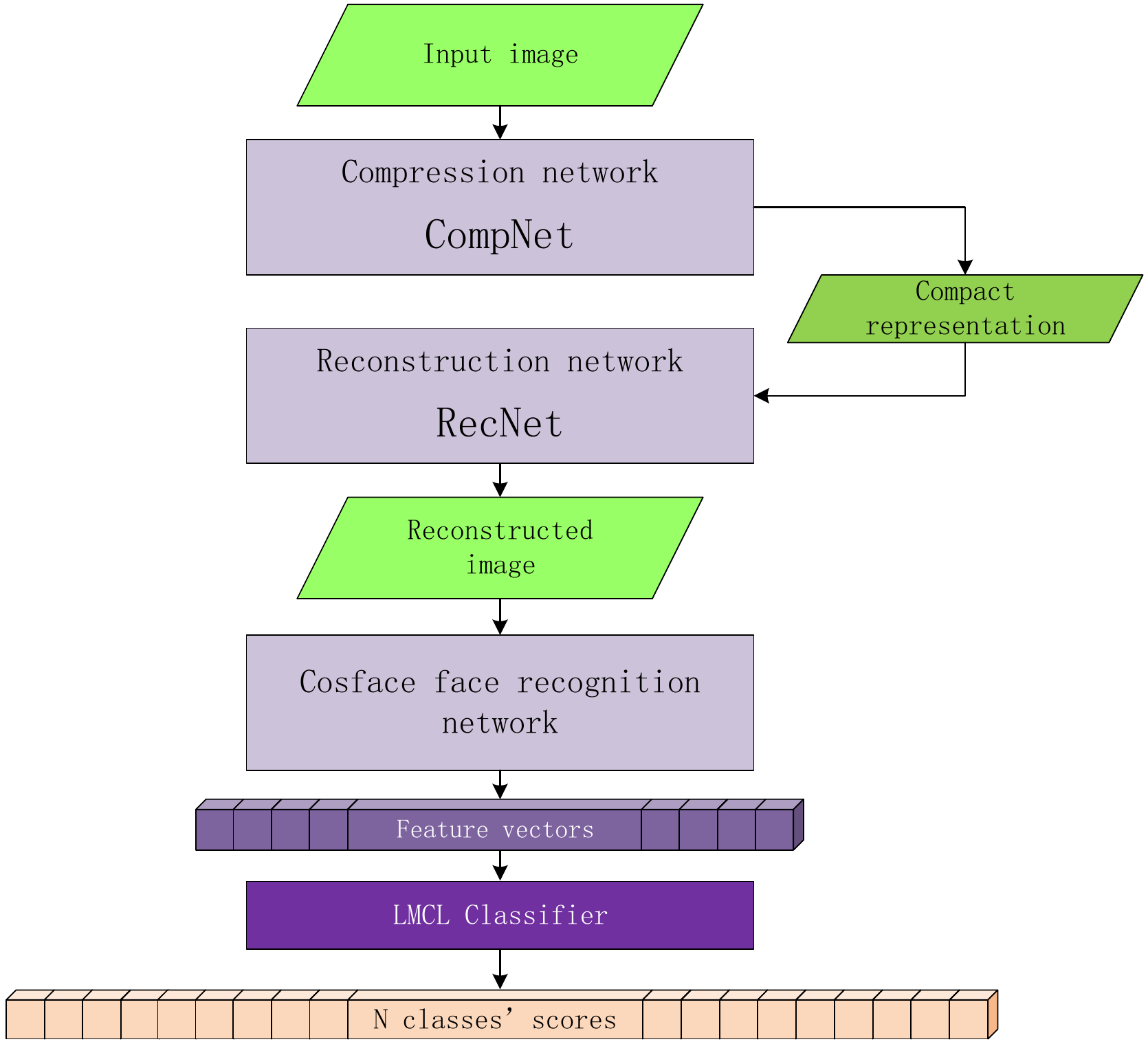}}
\caption{The training network of the overall framework.}
\label{fig7}
\end{figure}

During training process, our combined network omits the quantization and entropy coding for compact representation, due to the same reason mentioned already. As shown in Figure~\ref{fig7}, we directly input the compact representation of the output of CompNet to RecNet. Before the original image input into CompNet, we first normalize its pixel values from [0, 255] to [-1, 1].

The loss function of the combined network is still the loss function of the Cosface, that is LMCL. This is because we use the accuracy of face recognition as the only optimization goal and ignore the effect of the reconstructed images in joint training.

Since we have the pre-trained models and parameters of CompNet, RecNet, feature extraction part and classifier part of Cosface, they are all imported into the corresponding locations in the combined network. We use the same CASIA-WebFace dataset of the size of $112\times96$ as training set. Batchsize is set to 100, gradient descent algorithm is SGD, learning rate is 0.01 initially and then gradually reduced to 0.00001, thus we train it for 40 epochs.

\section{Experiment}
\subsection{Image recovery effect of Initially trained CompNet and RecNet}

\begin{figure}[htbp]
\centerline{\includegraphics[width=14.4cm]{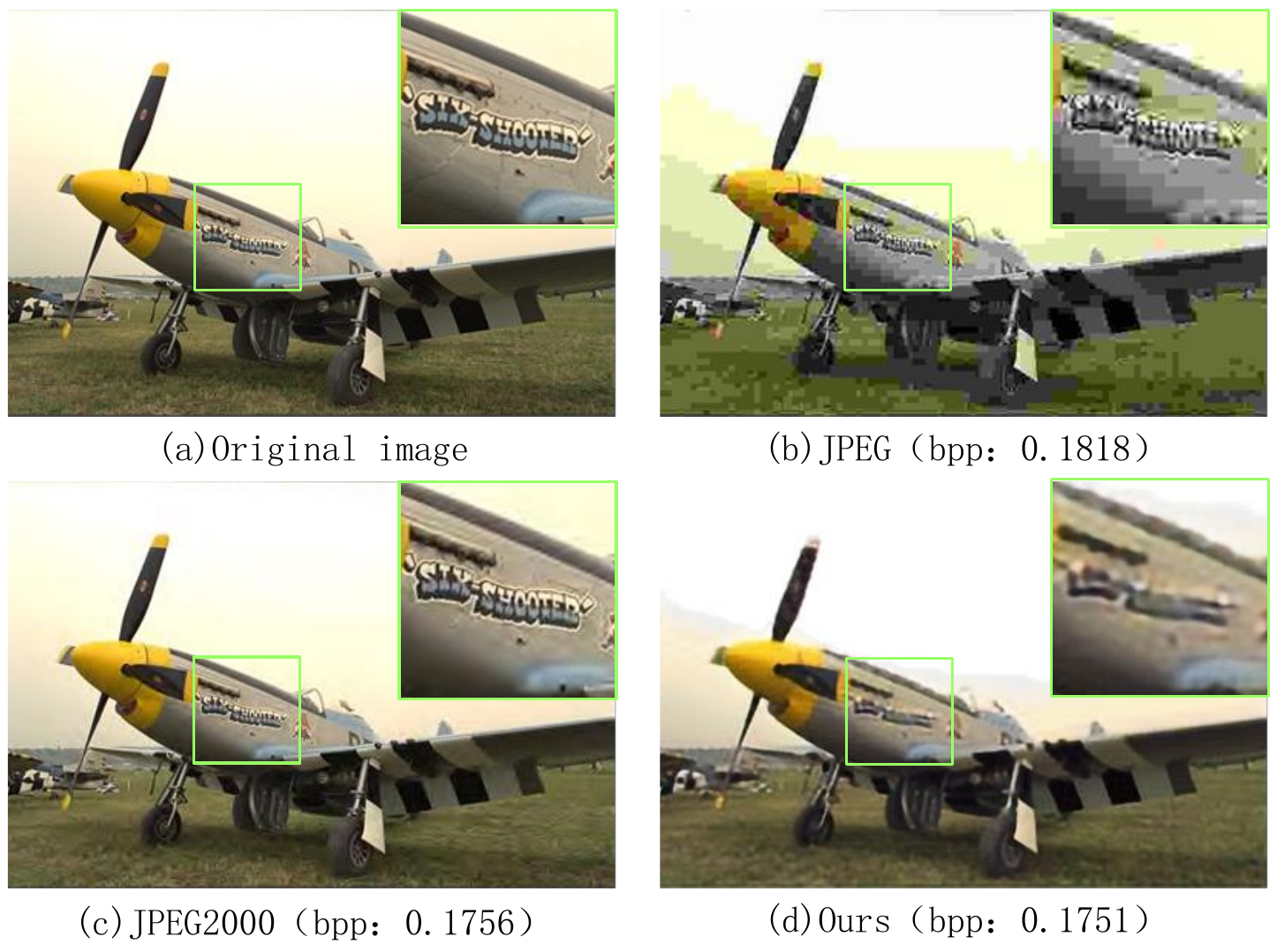}}
\caption{The effect of images restored by JPEG, JPEG2000 and our proposed network.}
\label{fig8}
\end{figure}

After completing the pre-training of CompNet and RecNet, we first use the dataset released by Kodak PhotoCD to roughly test its effect of image reconstruction. Figure~\ref{fig8} shows the image recovery effect of JPEG, JPEG2000, and our initially trained CompNet and RecNet at the same bits per pixel (bpp). Subjectively speaking, the images recovered by JPEG have serious blocking artifacts. Images compressed by the proposed CompNet and RecNet that we have initially trained are somewhat ambiguous. JPEG2000 compressed images have the best recovery effect among three methods.

However, the compression effect of different compression schemes at the same bit rates is not the main goal pursued in this paper. Therefore, we have not compared the effect of images restored by three compression schemes in detail by objective indicators (such as PSNR, SSIM). We mainly focus on the recognition accuracy when applying these compression schemes to the face recognition system. Therefore, after completing the joint training of our combined network proposed in this paper, we continue to focus on measuring the face verification accuracy.

\subsection{Comparison of face verification accuracy of LFW datasets compressed by different compression methods}
First, we used MTCNN to process face detection and calibration on the LFW face dataset, and generate 13233 face images of 5749 people with the picture size of 112$\times$96 as the benchmark test set. For this LFW\_112$\times$96 dataset, we respectively compress it by JPEG, JPEG2000 and the proposed CompNet that has been jointly trained with the face recognition network to the same size, and then test the face recognition accuracy on these datasets. (By the way, even with the smallest quality factor 0 when compressed by JPEG, the LFW\_112$\times$96 dataset can still not be compressed smaller than 10.8MB.) In addition, in order to evaluate the effect of joint optimization, we also test the verification accuracy of the LFW\_112$\times$96 face dataset compressed by our proposed compression network before joint optimization and the JPEG2000 scheme with the same compression ratio. The experimental results are shown in Table ~\ref{tab1}.

\begin{table}[htbp]
\caption{Varification accuracy on LFW\_112$\times$96 dataset compressed by different apporaches}
\begin{center}
  \begin{tabular}{ccc}
  \hline
  \textbf{Compression} & \textbf{Data} & \textbf{Varification}\\
  \textbf{Approach} & \textbf{Size} & \textbf{Accuracy}\\
  \hline
  Original LFW\_112$\times$96            & 30.6MB    & 99.28\%               \\
  CompNet after jointly trained   & 4.38MB    & 89.48\%               \\
  JPEG2000(Compression Ratio=88)  & 4.41MB    & 85.93\%               \\
  CompNet before jointly trained  & 2.79MB    & 75.87\%               \\
  JPEG(Quality Factor=0)          & 10.8MB    & 67.87\%               \\
  JPEG2000(Compression Ratio=134) & 2.78MB    & 64.85\%               \\
  \hline
\end{tabular}
\label{tab1}
\end{center}
\end{table}

Looking horizontally, the original LFW\_112$\times$96 dataset can achieve a face verification accuracy of 99.28\%. The size of the dataset compressed by CompNet after joint training becomes 1/7 of the original, and the face verification accuracy only decreases by 9.8\%, reaching 89.48\%, which is 3.55\% higher than the accuracy of 85.93\% of the JPEG2000 compressed dataset with the same compression ratio, and far higher than the poor accuracy of 67.87\% of the JPEG compressed dataset. In addition, when our pre-trained network (before jointly trained) compresses the LFW\_112$\times$96 dataset to 2.79MB (only 1/11 of the original dataset), the face verification accuracy is higher than the JPEG2000 method under the same compression ratio, and still higher than the JPEG method.

\begin{figure}[htbp]
\centerline{\includegraphics[width=9.6cm]{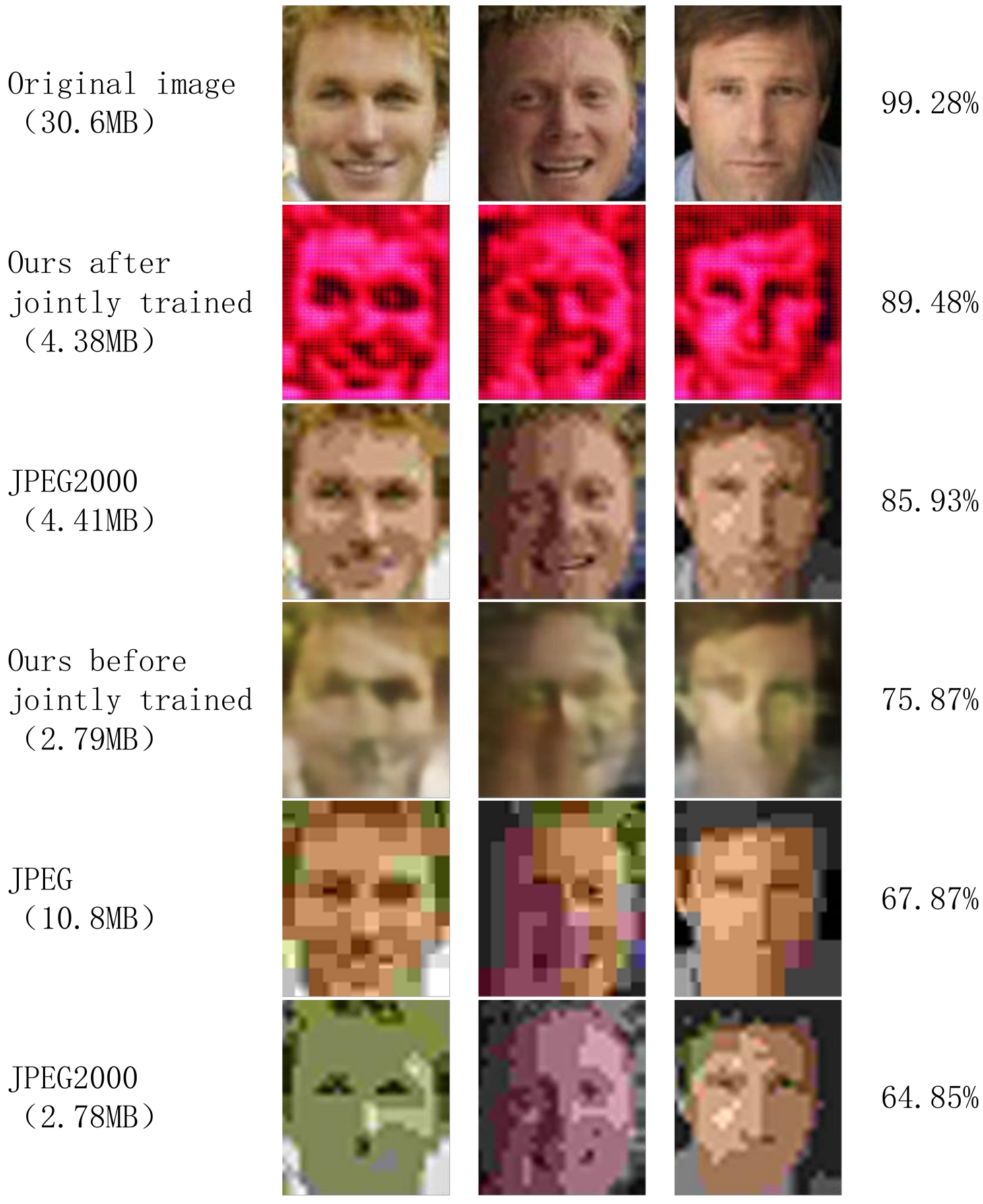}}
\caption{Some of the images in LFW\_112$\times$96 dataset compressed by different methods.}
\label{fig9}
\end{figure}

Looking vertically, joint training has significantly improved the face verification accuracy (from 75.87\% to 89.48\%), indicating that joint optimization aiming at improving face recognition accuracy is indeed effective.
For analysis, we present some of the face images in the LFW\_112$\times$96 dataset recovered by various compression schemes in Fig.~\ref{fig9}.

As can be seen from Fig.~\ref{fig9}, under the same compression ratio, the face images restored by the proposed network after joint training (4.38MB) are reddish and its restoration effect is worse than that compressed by JPEG2000 but get a higher accuracy. We hold the view that on the one hand, through joint training, the spatial structure of reconstructed image data may be adapted to face recognition, and the fact that images mainly retain information of channel R and abandon channel G and B indicates that it's grey information instead of color information that matters in face recognition; on the other hand, we have observed that in JPEG2000 compressed dataset (4.41MB) there are obvious discontinues color patches in the restored images. The abrupt boundary between these color patches is likely to be recognized by the feature extraction network as an ``edge'' feature, which will interfere the extraction of faces' essential features. The proposed jointly trained network (4.38MB) however does not have the above-mentioned color patches. Although it is reddish, it highlights the important parts of faces.

The face images recovered by the proposed network before jointly trained (2.79MB) is blurry, but still have higher accuracy than that recovered by JPEG (10.8MB) which have serious blocking artifacts and that recovered by JPEG2000 (2.78MB) which has much more serious discontinuous color patches. The reason can still be explained as that false boundaries cause by discontinuous blocking artifacts and color patches are affecting the extraction of essential face features.

In order to know the influence of the error caused by the quantization of the compact map on face recognition, we compare the verification accuracy of LFW dataset compressed by the proposed network before and after jointly trained with and without quantization. The results are shown in Table ~\ref{tab2}. It can be seen that without quantization, that is, the compact map that produced by CompNet is directly input into the RecNet and the face recognition network, the accuracy of the face verification is higher than that with quantization by 2\%--3\%. This is because there is error between the rounding results after quantization process and the original compact map's values. If we make some changes to the quantization process, there's still space for improvement in the face recognition effect.

\begin{table}[htbp]
\caption{Comparison of the varification on LFW dataset compressed by our network with and without quantization}
\begin{center}
\begin{tabular}{ccc}
\hline
 &\multicolumn{2}{c}{\textbf{Varification Accuracy}} \\
\textbf{Compression} & \multicolumn{2}{c}{\textbf{on LFW}} \\
\cline{2-3}
\textbf{Network} & \textbf{\textit{With}}& \textbf{\textit{Without}} \\
 & \textbf{\textit{quantization}} & \textbf{\textit{quantization}}\\
\hline
Our CompNet before jointly trained & 78.37\% & 75.87\%  \\
Our CompNet after jointly trained &	92.23\% & 89.48\%\  \\
\hline
\end{tabular}
\label{tab2}
\end{center}
\end{table}

\section{Conclusion}
In this paper, we propose a leaning based compression and reconstruction framework in the form of convolutional autoencoders, and combine it with a standard lossless image codec PNG. We connnect them with a existing face recognition network and trained them jointly. After experiment, we find that under the same compression ratio, the face verification accuracy of the LFW dataset compressed by our compression scheme is higher than that of the JPEG2000 scheme, and much higher than that of the JPEG scheme. In the future, we will continue to explore some continues functions to simulate the quantization process to futher improve the performance of face recognition.

\section*{Acknowledgment}

This work was supported by the National Natural Science Foundation of China under Grant No. 61474093 and by Tencent.

\bibliographystyle{unsrt}  


\end{document}